\documentclass[10pt,twocolumn,letterpaper]{article}

\usepackage{iccv}
\usepackage{times}
\usepackage{epsfig}
\usepackage{graphicx}
\usepackage{amsmath}
\usepackage{amssymb}
\usepackage[breaklinks=true,bookmarks=false]{hyperref}

\usepackage{booktabs}
\usepackage{dsfont}
\usepackage{multirow}
\usepackage[american]{babel}
\usepackage{subcaption}
\usepackage{xcolor}
\usepackage[capitalize]{cleveref}
\crefname{section}{Sec.}{Secs.}
\Crefname{section}{Section}{Sections}
\Crefname{table}{Table}{Tables}
\crefname{table}{Tab.}{Tabs.}

\iccvfinalcopy 

\def\iccvPaperID{****} 

\ificcvfinal\fi

\begin{document}

\title{ESTextSpotter: Towards Better Scene Text Spotting with Explicit Synergy in Transformer}

\author{
Mingxin Huang\textsuperscript{1}$^\dag$
\quad Jiaxin Zhang\textsuperscript{2}$^\dag$
\quad Dezhi Peng\textsuperscript{1}
\quad Hao Lu\textsuperscript{3}
\quad Can Huang\textsuperscript{2}
\\
\quad Yuliang Liu\textsuperscript{3}
\quad Xiang Bai\textsuperscript{3}
\quad Lianwen Jin\textsuperscript{1}$^*$
\\
\textsuperscript{1}{South China University of Technology} 
\quad \quad \textsuperscript{2}{ByteDance}
\\
\textsuperscript{3}{Huazhong University of Science and Technology}
\\
{\tt\small eelwjin@scut.edu.cn}
}

\maketitle
\ificcvfinal\thispagestyle{empty}\fi

\begin{abstract}
In recent years, end-to-end scene text spotting approaches are evolving to the Transformer-based framework. While previous studies have shown the crucial importance of the intrinsic synergy between text detection and recognition, recent advances in Transformer-based methods usually adopt an implicit synergy strategy with shared query, which can not fully realize the potential of these two interactive tasks. In this paper, we argue that the explicit synergy considering distinct characteristics of text detection and recognition can significantly improve the performance text spotting. To this end, we introduce a new model named Explicit Synergy-based Text Spotting Transformer framework (ESTextSpotter), which achieves explicit synergy by modeling discriminative and interactive features for text detection and recognition within a single decoder. Specifically, we decompose the conventional shared query into task-aware queries for text polygon and content, respectively. Through the decoder with the proposed vision-language communication module, the queries interact with each other in an explicit manner while preserving discriminative patterns of text detection and recognition, thus improving performance significantly. Additionally, we propose a task-aware query initialization scheme to ensure stable training. Experimental results demonstrate that our model significantly outperforms previous state-of-the-art methods. Code is available at \url{https://github.com/mxin262/ESTextSpotter}.
\end{abstract}
\let\thefootnote\relax\footnotetext{$^\dag$Equal contribution.}
\let\thefootnote\relax\footnotetext{$^*$Corresponding author.}

\section{Introduction}
\label{sec:intro}

End-to-end text spotting, aiming at building a unified framework for text detection and recognition in natural scenes, has received great attention in recent years~\cite{liu2018fots,liao2019mask,liu2020abcnet}. Intuitively, the position and shape of the text in the detection can help the text recognition accurately extract the content of the text. Similarly, the position and classification information in recognition can also guide the detector to distinguish between different text instances and the background. Such mutual interaction and cooperation between text detection and recognition are recently known as scene text spotting synergy~\cite{huang2022swintextspotter}, which aims to produce a combined effect greater than the sum of their separate effects. Indeed, synergy is the key to the success in literature.

\begin{figure}[t!]
    \centering
    
    \includegraphics[width=\linewidth]{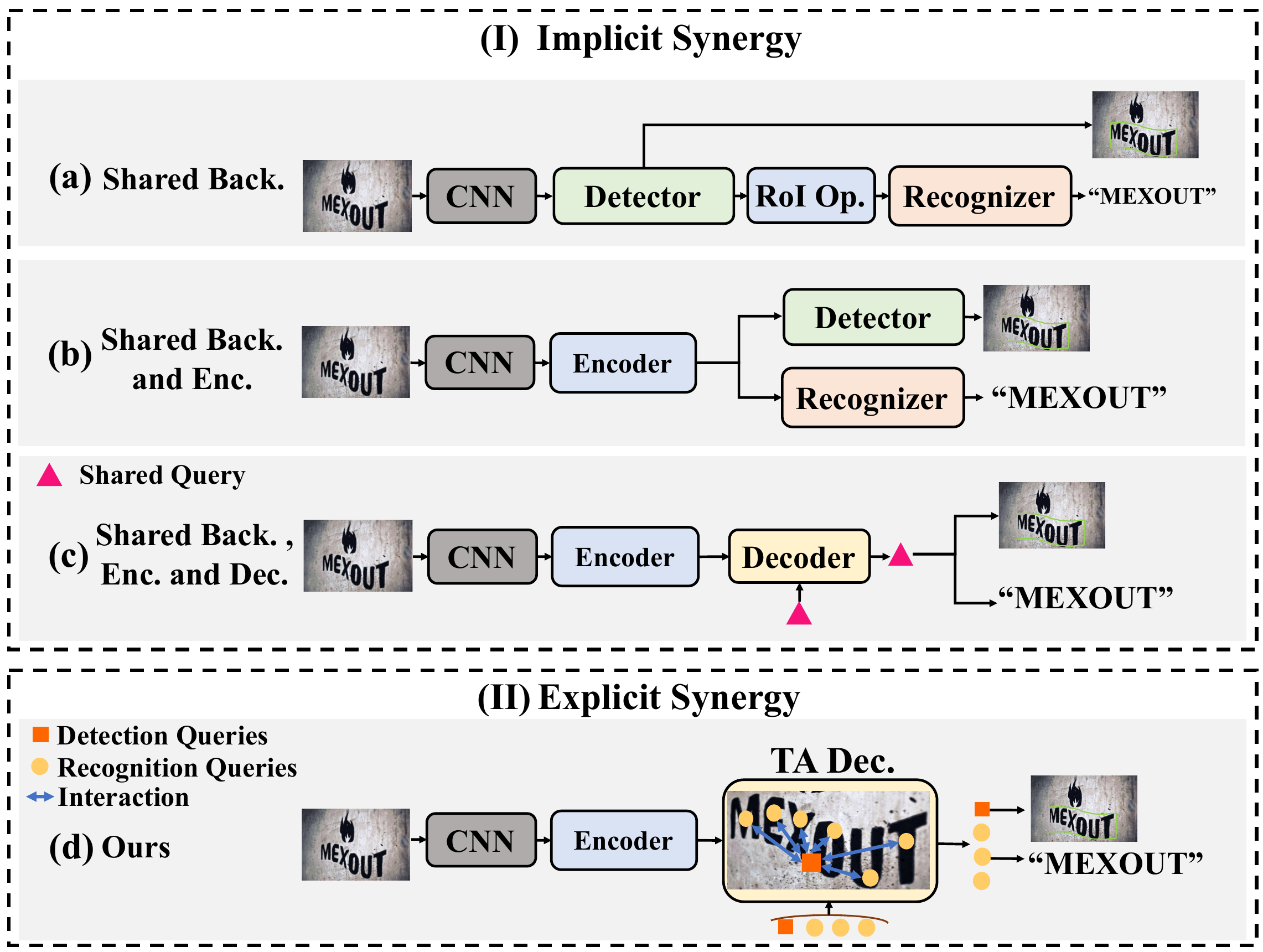}
    \caption{Comparison of implicit and explicit synergy between text detection and recognition. Implicit synergy is achieved by sharing parameters and features. Explicit synergy is attained by explicitly modeling discriminative and interactive features. Back.: backbone. Enc. (Dec.): encoder (decoder). TA Dec.: task-aware decoder.
    }
    \label{fig:intro}
\end{figure}

In the past few years, many methods attempt to join text detection and recognition by proposing a new Region-of-Interest (RoI) operation to achieve the synergy between text detection and text recognition~\cite{liu2018fots,feng2019textdragon,wang2020all,liu2020abcnet,wang2021pgnet}, as shown in Figure~\ref{fig:intro}(a). They follow the classical two-stage pipeline, which first locates the text instance and then extracts the text content in the corresponding region of interest (RoI). However, the interaction between detection and recognition is insufficient through sharing a backbone, as observed in recent research~\cite{huang2022swintextspotter}. A recent study, TESTR~\cite{zhang2022text}, develops dual-decoder framework to further share an encoder, but there is still a lack of interaction between the two tasks, as presented in Figure~\ref{fig:intro}(b).
Therefore, some researchers~\cite{kittenplon2022towards,ye2022deepsolo} begin to explore better synergy based on the Transformer~\cite{vaswani2017attention} architecture. For instance, TTS~\cite{kittenplon2022towards} takes a step toward unifying the detector and recognizer into a single decoder with shared query for both two tasks as illustrated in Figure~\ref{fig:intro}(c). DeepSolo~\cite{ye2022deepsolo} further adopts a group of shared queries to encode the characteristics of text. Although these approaches~\cite{kittenplon2022towards,ye2022deepsolo} develop a more concise and unified framework, they fail to consider distinct feature patterns of these two tasks. We formulate the above-mentioned methods as utilizing an implicit synergy that shares parameters and features between the detection and recognition, but lacks explicit modeling between them, as shown in Figure~\ref{fig:intro}(I). The full potential of two tasks can not be realized by implicit synergy alone without considering the unique characteristics of each task~\cite{song2020revisiting,wu2020rethinking}. For instance, while DeepSolo has demonstrated promising end-to-end results on Total-Text~\cite{ch2020total}, its detection performance falls short of that achieved by the dedicated detection method~\cite{tang2022few}.

In this paper, we propose an Explicit synergy Text Spotting
Transformer framework, termed ESTextSpotter, stepping toward explicit synergy between text detection and recognition. Compared to previous implicit synergy, ESTextSpotter explicitly models discriminative and interactive features for text detection and recognition within a single decoder, as illustrated in Figure~\ref{fig:intro}(d). Typically, we design a set of task-aware queries to model the different feature patterns of text detection and recognition, which include detection queries encoding the position and shape information of the text instance, and recognition queries encoding the position and semantics information of the character. The position information of the character is obtained through an attention mechanism similar to previous works~\cite{fang2021read,wang2021two}. Then, detection queries and recognition queries are sent into a task-aware decoder that is equipped with a vision-language communication module to enhance the explicit interaction. Previous works~\cite{zhang2022text,kittenplon2022towards,ye2022deepsolo} have used learnable embeddings to initialize the queries. However, these randomly initialized parameters will disrupt the training of the vision-language communication module. Therefore, we propose a task-aware queries initialization (TAQI) to promote stable training of the vision-language communication module. Besides, inspired by ~\cite{li2022dn,zhang2022dino}, we also employ a denoising training strategy to expedite convergence.

Extensive experiments demonstrate the effectiveness of our method: 1) For text detection, ESTextSpotter significantly outperforms previous detection methods by an average of $3.0\%$ in terms of the H-mean on two arbitrarily-shaped text datasets, $1.8\%$ on two multi-oriented datasets, and $3.0\%$ on Chinese and multi-lingual datasets; 
2) For English text spotting, ESTextSpotter consistently outperforms previous methods by large margins;
3) ESTextSpotter also significantly outperforms previous methods on multilingual text spotting including Chinese text (ReCTS), African Amharic text (HUST-Art), and Vietnamese text (VinText), with an average of $4.8\%$ in terms of the end-to-end H-mean. 

In conclusion, our contributions can be summarized as follows.
\begin{itemize}
\item 
We introduce ESTextSpotter, a simple yet efficient Transformer-based approach for text spotting that adopts task-aware queries within a single decoder, which allows it to effectively realize explicit synergy of text detection and recognition, thereby unleashing the potential of these two tasks.

\item 
We propose a vision-language communication module designed to enhance explicit synergy, which utilizes a novel collaborative cross-modal interaction between text detection and recognition. Moreover, we introduce a task-aware query initialization module to guarantee stable training of the module.

\item 
We achieve significant improvements over state-of-the-art methods across eight challenging scene text spotting benchmarks.
\end{itemize}

\section{Related Work}
\label{sec:rela}

\begin{figure*}[!thp]
    \centering
    \includegraphics[width=\textwidth]{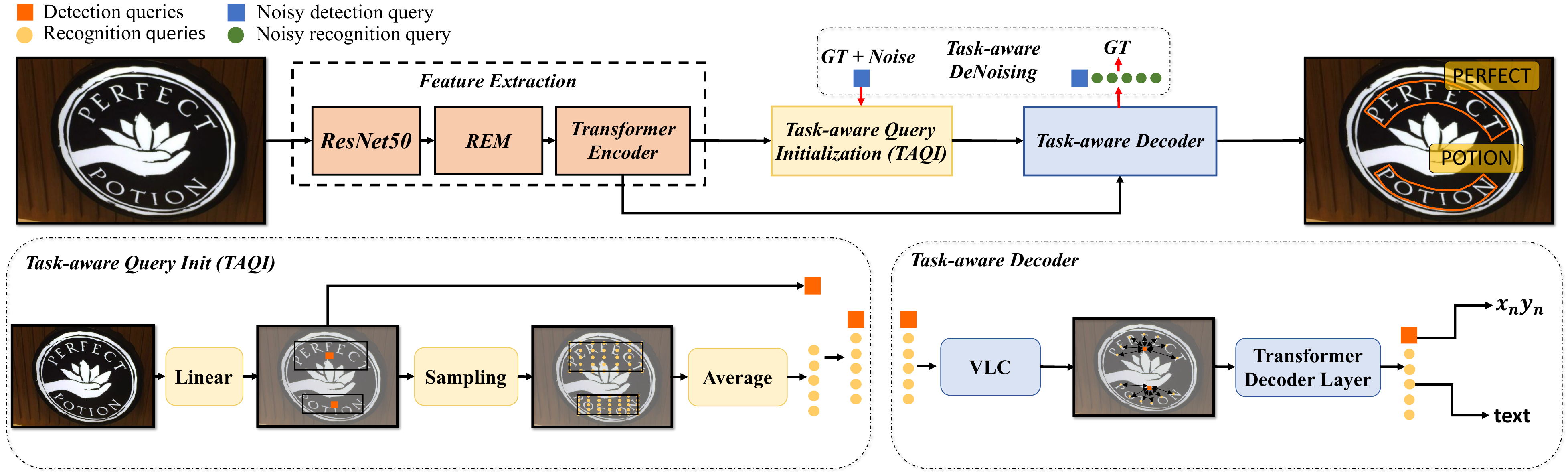}\vspace{-10pt}
    \caption{
        The framework of the proposed ESTextSpotter. The image features are extracted in the feature extraction process. Then the Task-aware Query Initialization is used to generate the task-aware queries including detection and recognition queries. Then the task-aware queries are sent into the task-aware decoder to obtain the detection and recognition results simultaneously. REM is the
        Receptive Enhancement Module. VLC means the vision-language communication module. The red arrow means only used in the training stage.
    }
    \label{fig:frame}
\end{figure*}

\paragraph{End-to-End Scene Text Spotting.} Classical methods~\cite{wang2011end,jaderberg2016reading,liao2018textboxes++} have some limitations in addressing scene text spotting, such as error accumulation, sub-optimization, and low inference efficiency. To overcome these problems, a paradigm shift has been witnessed from shallow learning to end-to-end learning. In particular, Li \etal~\cite{li2017towards} integrated detection and recognition into a unified end-to-end framework.
However, this method mainly handles horizontal texts. Some researchers introduced special RoI operations, such as Text-Align~\cite{he2018end} and RoI-Rotate~\cite{liu2018fots}, to sample the oriented text features into regular ones for text recognition.
Liao \etal~\cite{lyu2018mask} proposed Mask TextSpotter, which introduces a character segmentation module to take the advantage of character-level annotations, to solve the problem of arbitrarily-shaped scene text spotting. TextDragon~\cite{feng2019textdragon} further proposed RoISlide to fuse features from the predicted segments for text recognition. Qin \etal~\cite{qin2019towards} proposed RoI Masking to suppress the background noise by multiplying segmentation masks with features. Wang \etal~\cite{wang2020all} used a boundary detector to cooperate with the use of Thin-Plate-Spline (TPS)~\cite{bookstein1989principal}. Mask TextSpotter v3~\cite{liao2020mask} proposed a Segmentation Proposal Network (SPN) to generate accurate proposals for arbitrarily-shaped text. MANGO~\cite{qiao2021mango} developed a Mask Attention module to coarsely localize characters, which requires character-level annotations. ABCNet~\cite{liu2020abcnet} and its improved version ABCNet v2~\cite{liu2021abcnetv2} used the parametric bezier curve to model the curved text and developed Bezier-Align for rectifying curved text. The methods discussed above mainly focus on designing shape-aware Region of Interest (RoI) sampling, while merely achieving synergy by sharing the backbone.

\vspace{-10pt}
\paragraph{Text Spotting Transformer.} To further enhance the interaction between detection and recognition, TETSR~\cite{zhang2022text} developed a dual-decoder framework to share both backbone and encoder between two tasks, and only detection and recognition head are isolated. SwinTextSpotter~\cite{huang2022swintextspotter} further proposed a Recognition Conversion to implicitly guide the recognition head through incorporating the detection and back-propagate recognition information to the detector. TTS~\cite{kittenplon2022towards} attempted to unify the detector and recognizer in a single decoder using a shared query. To encode the characteristics of text in the queries, DeepSolo~\cite{ye2022deepsolo} utilized a group of point queries based on the center line. Similarly, SPTS~\cite{peng2022spts} adopted an auto-regressive framework~\cite{chen2021pix2seq} that most parameters are shared between text detection and recognition.

Although the Transformer has shown great potential in text spotting, current methods still have limitations. Firstly, the dual-decoder framework~\cite{zhang2022text} lacks interaction between text detection and recognition, which limits the performance. Secondly, the shared query in the single decoder framework~\cite{kittenplon2022towards,ye2022deepsolo} does not fully consider the distinct feature patterns of these two tasks. Note that, while a closely related work, SwinTextSpotter~\cite{huang2022swintextspotter}, also attempts to explore the synergy between text detection and recognition, it does not fully achieve explicit synergy. This is because it back-propagates recognition information to the detector without explicitly modeling the relationship between text detection and recognition.

\section{
Methodology
}
In this paper, we propose an Explicit synergy-based Text Spotting Transformer framework, termed ESTextSpotter. The key idea of ESTextSpotter is to explicitly model discriminative and interactive features for text detection and recognition within a single decoder. The overall architecture is shown in Figure~\ref{fig:frame}. After obtaining image features through the feature extraction process consisting of ResNet50, receptive enhancement module (REM), and Transformer encoder, task-aware queries are generated using the Task-aware Query Initialization module (TAQI), which includes detection and recognition queries. These queries are then sent into the task-aware decoder to explicitly model discriminative and interactive features for text detection and recognition simultaneously. During training, inspired by previous works~\cite{li2022dn, zhang2022dino}, we utilize a task-aware DeNoising training strategy to accelerate convergence. Detailed implementations will be provided in the following subsections.

\vspace{-2pt}
\subsection{Receptive Enhancement Module} 

Following previous works~\cite{liao2019mask,liu2021abcnetv2,zhang2022text}, we adopt ResNet50~\cite{he2016deep} as our backbone. To enhance the receptive field of the features, we send the feature map $res_5$ output from the ResNet50 to the receptive enhancement module (REM), which uses a convolutional layer to downsample the feature map. Then, we send the output of the REM, as well as the feature maps $res_3$ to $res_5$, to the Transformer encoder~\cite{zhu2020deformable} to model long-range dependencies across various scales. Finally, the output of the Transformer encoder is fed into the subsequent modules.

\vspace{-2pt}
\subsection{Task-aware Query Initialization}
\label{TAQI}
\vspace{-2pt}

Previous works~\cite{zhang2022text, kittenplon2022towards, ye2022dptext, ye2022deepsolo} have utilized learnable embeddings to initialize the queries. However, these randomly initialized parameters will disrupt the training of the vision-language communication module. Therefore, we propose task-aware query initialization (TAQI) to improve the stability of the vision-language communication module during training. Firstly, we use a linear layer to generate text classification scores from the output of the Transformer encoder. Then, we select the top $N$ features based on the text classification scores, and these features are sent into a linear layer to initialize proposals for detection queries. For recognition queries, we sample the features $\mathbf{F_p} \in \mathbb{R}^{N \times H \times T \times C}$ from the proposals and average over the height dimension to initialize recognition queries. Here, $N$ represents the maximum number of predictions identified in DETR~\cite{carion2020end}, while $T$ represents the length of recognition queries, and $C$ represents the feature dimension. Benefiting from the decomposition of the conventional shared query, TAQI encodes the boundary and content information into the detection and recognition queries, respectively.

\subsection{Task-aware Decoder}
\label{ESTextSpotter decoder}

\begin{figure}[!t]
    \centering
    \includegraphics[width=\linewidth]{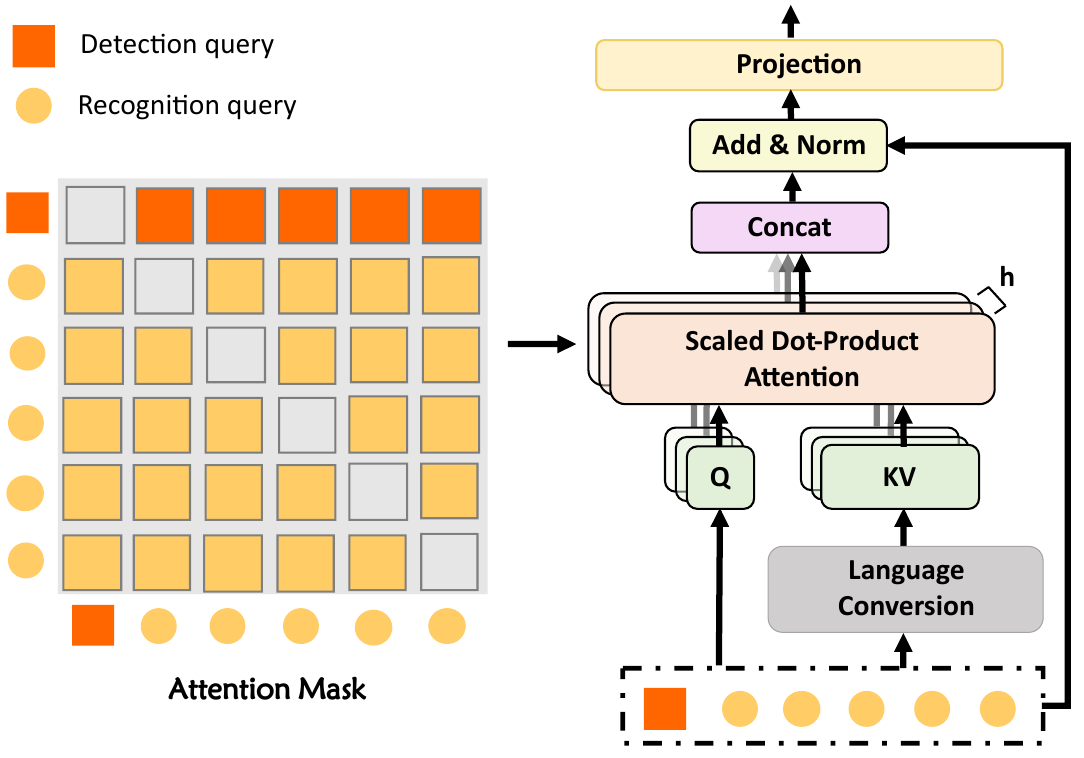}
    \caption{
        \textbf{Illustration of the Vision-language communication module}. The language conversion extracts the semantic features in recognition queries. Then the visual and semantic information can interact in a cross-modal perspective. h is the number of parallel attention heads.
    }
    \label{fig:m_attn_with_init}
\end{figure}

After obtaining the task-aware queries $\mathbf{S} \in \mathbb{R}^{N \times (T+1) \times C}$, including detection queries $\mathbf{G} \in \mathbb{R}^{N \times C} $ and recognition queries $\mathbf{R} \in \mathbb{R}^{N \times T \times C}$, they are sent to the task-aware decoder to interact and mutually promote each other. We first enhance the explicit synergy between task-aware queries from a cross-modal perspective in the proposed vision-language communication module as illustrated in Figure~\ref{fig:m_attn_with_init}. A Language Conversion is designed to extract semantic features in recognition queries and map them into language vectors $\mathbf{L} \in \mathbb{L}^{N \times (T+1) \times C}$, which is defined as follows:
\begin{equation}\small
\mathbf{P} = {\tt softmax}(\mathbf{W_1 R})\,,
\end{equation}
\begin{equation}\small
\mathbf{L} = {\tt cat}(\mathbf{G}, \mathbf{W_2 P})\,,
\end{equation}
where $\mathbf{W_1} \in \mathbb{R}^{C \times U}$ and $\mathbf{W_2} \in \mathbb{R}^{U \times C}$ are trainable weights. $U$ indicates the character class number. ${\tt cat}$ is the concatenation operation. Then the task-aware queries $\mathbf{S}$ and language vectors $\mathbf{L}$ are sent to a vision-language attention module, which is formalized as:

\begin{equation}\small
\mathbf{M}_{ij} = \left\{
\begin{aligned}
0 & , & i \ne j\,, \\
-\infty & , & i=j\,.
\end{aligned}
\right.
\end{equation}
\begin{equation}\small
\mathbf{F} = {\tt softmax}(\frac{(\mathbf{S} + {\rm PE}(\mathbf{S}))(\mathbf{L} + {\rm PE}(\mathbf{L}))^T}{\sqrt{D}} + \mathbf{M})\mathbf{L}\,.
\end{equation}
$\rm PE$ indicates the position encoding used by DETR~\cite{carion2020end}. The attention mask $\mathbf{M}$ is designed to prevent the queries from over-focusing ``itself''.

After exchanging vision-language information, we send the task-aware queries to the Transformer decoder layer. Consistent with prior research works~\cite{zhang2022text,ye2022dptext,ye2022deepsolo}, we first initially incorporate an intra-group self-attention module to enhance the relationship between task-aware queries within one text instance. Subsequently, we use an inter-group self-attention module to model the relationship between distinct text instances. The outputs of these modules are then fed into a multi-scale deformable cross-attention module to extract the text features from the Transformer encoder. Finally, we employ two linear layers to predict the detection and recognition results.

During the decoding stage, the detection queries extract the position and shape information of text instances, while the recognition queries comprise the semantic and positional information of characters. When explicitly modeling the relationship in the task-aware decoder, the positional information of the character available in the recognition queries can assist the detection queries in accurately locating the text. Similarly, the positional and shape information of the text instance present in the detection queries can help the recognition queries in extracting character features. As a result, this explicit synergy between detection and recognition queries unleashes the potential of both text detection and recognition while also preserving the distinct feature patterns in the detection and recognition queries.

\vspace{-5pt}
\paragraph{Detection and Recognition Format.} 
In contrast to previous works that utilize serial point queries to model the curve~\cite{zhang2022text, ye2022dptext, ye2022deepsolo} or freeze the model weights to train a segmentation head~\cite{kittenplon2022towards}, we develop a simpler approach to generate the detection results. We send the detection queries into two feed-forward layers, in which one predicts the proposals $ (x, y, h, w) $, while the other predicts $Z$ polygon offsets $ (\Delta x_1, \Delta y_1, \Delta x_2, \Delta y_2, ... ,\Delta x_Z, \Delta y_Z) $ based on the center point $(x, y)$ of each proposal. Z is 16. Reconstructing the polygon can be formulated as:
\begin{equation}\small
x_i = x + \Delta x_i\,,
\end{equation}
\begin{equation}\small
y_i = y + \Delta y_i\,.
\end{equation}
In this way, we can predict the detection result directly through detection queries without serial control points or freezing the model weights to train a segmentation head. Following the decoding process, the recognition queries can efficiently extract character features. We utilize a linear layer to convert the recognition queries into characters, similar to ~\cite{zhang2022text}.

\subsection{Optimization}

\paragraph{Task-aware DeNoising.} Recently, some researchers~\cite{li2022dn,zhang2022dino} propose the DeNoising training to accelerate the convergence of the DETR~\cite{carion2020end}. However, these methods are specifically designed for detection. Therefore, we develop a Task-aware DeNoising (TADN) strategy for text spotting to accelerate the convergence, as presented in Figure~\ref{fig:frame}. Following previous works~\cite{li2022dn,zhang2022dino}, we add center shifting and box scaling in ground truth boxes, termed noise boxes. The noise boxes are transformed into noise detection queries by linear layers, and the noise recognition queries are initialized by TAQI. The noise detection and recognition queries are concatenated and sent to the task-aware decoder, which is responsible for reconstructing the ground truth boxes and obtaining the corresponding recognition results. TADN more focuses on text spotting rather than detection, as opposed to previous denoising training methods~\cite{li2022dn,zhang2022dino}.

\paragraph{Loss.} The training process of ESTextSpotter is a set prediction problem that uses a fixed number of outputs to match the ground truths. Inspired by the DETR-like methods~\cite{carion2020end,zhu2020deformable,liu2022dabdetr}, we utilize the Hungarian algorithm~\cite{kuhn1955hungarian} to perform pairwise matching and minimize the prediction-ground truth matching cost $\mathcal{C}_{match}$ as:
\begin{equation}\small
\hat \sigma = \underset{\sigma}{\arg \min} \sum_{i=1}^{N}\mathcal{C}_{match}(Y_i, \hat Y_{\sigma(i)})\,,
\end{equation}
where $Y_i$ is the ground truth and $\hat Y_{\sigma(i)}$ is the prediction. $N$ is the number of the predictions indexed by $\sigma(i)$. The cost function $\mathcal{C}_{match}$ is defined as:
\begin{equation}\small
\mathcal{C}_{match}(Y_i, \hat Y_{\sigma(i)}) = \lambda_{c}\mathcal{C}_{c}(\hat p_{\sigma(i)}(c_i)) + \mathds{1}_{\left\{c_i \neq \emptyset\right\}} \lambda_{b}\mathcal{C}_{b}(b_i, \hat b_{\sigma(i)})\,,
\end{equation}
where $c_i$ and $b_i$ are the ground truth class and bounding box, and $\hat b_{\sigma(i)}$ represents the prediction of bounding box. $\hat p_{\sigma(i)}(c_i)$ is the probability of prediction for class $c_i$. $\lambda_{c}$ and $\lambda_{b}$ are the weights for the classification and bounding box. After the Hungarian algorithm, the prediction and ground truth can be one-to-one matched. The training loss is as follows:
\begin{multline}\small
    \mathcal{L}(Y_i, \hat Y_{\sigma(i)}) = \alpha_{c}\mathcal{L}_{c}(\hat p_{\sigma(i)}(c_i)) + \mathds{1}_{\left\{c_i \neq \emptyset\right\}} \mathcal{L}_{b}(b_i, \hat b_{\sigma(i)}) \\
     + \mathds{1}_{\left\{c_i \neq \emptyset\right\}} \alpha_{r}\mathcal{L}_{r}(r_i, \hat r_{\sigma(i)}) + \mathds{1}_{\left\{c_i \neq \emptyset\right\}} \alpha_{p}\mathcal{L}_{p}(p_i, \hat p_{\sigma(i)})\,,
\end{multline}
where $\alpha_{c}$, $\alpha_{b}$, $\alpha_{p}$, and $\alpha_{r}$ are the loss weights for the classification, bounding box, polygon, and recognition, respectively. The classification loss $\mathcal{L}_{c}$ is the focal loss~\cite{lin2017focal}. The bounding box loss $\mathcal{L}_{b}$ consists of the 
$\ell_1$ loss and the GIoU loss~\cite{rezatofighi2019generalized}. The polygon loss uses the $\ell_1$ loss as well. The recognition loss is the standard cross entropy loss.

\section{Experiments and Results}
We conduct experiments on common benchmarks to evaluate ESTextSpotter, including multi-oriented ICDAR2015~\cite{karatzas2015icdar} and MSRA-TD500~\cite{yao2012detecting},  multilingual datasets
ReCTS~\cite{zhang2019icdar}, Vintext~\cite{m_Nguyen-etal-CVPR21}, HUST-ART~\cite{dikubab2022comprehensive}, and ICDAR2019-MLT~\cite{nayef2019icdar2019}, arbitrarily shaped datasets TotalText~\cite{ch2020total}, and SCUT-CTW1500~\cite{liu2019curved}.

\begin{table*}[!t]
  \centering
  \newcommand{\tabincell}[2]{\begin{tabular}{@{}#1@{}}#2\end{tabular}}
\footnotesize
\caption{Detection results on the Total-Text, SCUT-CTW1500, MSRA-TD500, and ICDAR 2015 datasets. AAvg. means the average in arbitrarily-shaped text. MAvg. means the average in multi-oriented text.
Bold indicates SOTA, and underline indicates the second best.}
\resizebox{\linewidth}{!}{%
  \begin{tabular}{@{}l ccc ccc ccc ccc ccc c c@{}}
    \toprule
    \multirow{2}*{Methods}  & \multicolumn{3}{c}{Total-Text} & \multicolumn{3}{c}{SCUT-CTW1500} &  \multicolumn{3}{c}{MSRA-TD500}&
    \multicolumn{3}{c}{ICDAR 2015} & \multicolumn{3}{c}{ReCTS} &
    \multirow{2}*{AAvg.} & \multirow{2}*{MAvg.} \\
    \cmidrule(r){2-4} \cmidrule(r){5-7} \cmidrule(r){8-10} \cmidrule(r){11-13} \cmidrule(r){14-16}
        & R  & P & H & R  & P & H & R  & P & H &R&P&H &R&P&H & & \\
    \midrule
    TextDragon \cite{feng2019textdragon}  & 75.7  & 85.6  & 80.3  & 82.8  & 84.5  & 83.6  &  -- & --  & -- &--&--&--  &--&--&-- & 82.0 & --\\
    PSENet--1s \cite{wang2019shape}  & 78.0 & 84.0 & 80.9 & 79.7 & 84.8 & 82.2 & -- & -- & --  &85.5&88.7&87.1  &83.9&87.3&85.6  & 81.6 & --\\
    CRAFT \cite{baek2019character} & 79.9 & 87.6 & 83.6 & 81.1 & 86.0 & 83.5 & 78.2 & 88.2 & 82.9 &84.3&89.8&86.9  &--&--&-- & 83.6 & 84.9\\
    PAN \cite{wang2019efficient} & 81.0 & 89.3 & 85.0 & 81.2 & 86.4 & 83.7 &  83.8 & 84.4 & 84.1 &81.9&84.0&82.9  &--&--&-- & 84.4 & 83.5\\
    DBNet \cite{liao2020real} &82.5&87.1&84.7  &80.2&86.9&83.4  &77.7&76.6&81.9 &82.7&88.2&85.4 &--&--&-- & 84.1 & 83.7\\
    DRRG \cite{zhang2020deep}  & 84.9 & 86.5 & 85.7 & 83.0 &  85.9 & 84.5 & 82.3 & 88.1 & 85.1 &84.7&88.5&86.6  &--&--&-- & 85.1 & 85.9\\
    CounterNet \cite{wang2020contournet} & 83.9 & 86.9 & 85.4 & 84.1 & 83.7 & 83.9 & -- & -- & -- &86.1&87.6&86.9 &--&--&-- & 84.7 & --\\
    FCENet \cite{zhu2021fourier} & 82.5 & 89.3 & 85.8 & 83.4 & 87.6 & 85.5 & -- & -- & -- &82.6 & 90.1 & 86.2 &--&--&-- & 85.7 & --\\
    PCR \cite{dai2021progressive} & 82.0 & 88.5 & 85.2 & 82.3 & 87.2 & 84.7 & 83.5 & 90.8 & 87.0 & -- & -- & -- &--&--&-- & 85.0 & --\\
    MOST \cite{he2021most} & --  & --  & -- & -- & -- & -- & 82.7 & 90.4 & 86.4 & 87.3 & 89.1 & 88.2 &--&--&-- & -- & 87.3\\
    TextBPN\cite{zhang2021adaptive} & 85.2 & 90.7 & 87.9  & 83.6 & 86.5 & 85.0 &84.5 & 86.6 & 85.6 &-- & -- & -- &-- & -- & -- & 86.5 & -- \\ 
    ABCNet v2\cite{liu2021abcnetv2} & 84.1 & 90.2 & 87.0 & 83.8 & 85.6 & 84.7 &81.3& 89.4 & 85.2 &86.0&90.4& 88.1 & \underline{87.5} & \underline{93.6} & \underline{90.4} &85.9 & 86.7\\ 
    PAN++\cite{liu2021abcnetv2} & 81.0 & 89.9 & 85.3 & 81.1 & 87.1 & 84.0 & \underline{85.6} & 91.4 & \underline{88.4} & 83.9 & 91.4  & 87.5 & -- & -- & -- & 84.7 & 88.0\\
    DBNet++\cite{liao2022real}  &83.2 &88.9  &86.0  &  82.8 &87.9 & 85.3 &83.3 &91.5 &87.2 &83.9 &90.9  &87.3 &-- & -- & --& 85.7 & 87.3 \\ 
    FSGNet\cite{tang2022few} & 85.7 & 90.7 & \underline{88.1}  & 82.4 & 88.1 & 85.2 &84.8 & \underline{91.6} & 88.1 &86.7 & 91.1 & 88.8 &-- & -- & --& 86.7 & \underline{88.5} \\ 
    TESTR\cite{zhang2022text} & 81.4 & \textbf{93.4}  & 86.9 & 82.6 & \textbf{92.0} & \underline{87.1} &-- & -- & -- &\textbf{89.7} & 90.3  & \underline{90.0} &-- & -- & -- & \underline{87.0} & -- \\ 
    DeepSolo\cite{ye2022deepsolo} & 82.1 & \underline{93.1}  & 87.3 & -- & -- & -- & -- & -- & -- & 87.4 & 92.8 & 90.0 &-- & -- & -- & -- & -- \\ 
    \hline
    ESTextSpotter-Polygon (Ours) & \textbf{88.1} & 92.0  & \textbf{90.0} & \textbf{88.6}  & \underline{91.5} & \textbf{90.0} & \textbf{86.3} & \textbf{92.9} & \textbf{89.5} & \underline{89.6} & \textbf{92.5} & \textbf{91.0} & \textbf{91.3} & \textbf{94.1} & \textbf{92.7} & \textbf{90.0} & \textbf{90.3} \\
    \bottomrule
  \end{tabular}
}
\label{tab:cp_sota_det}
\end{table*}

\begin{table*}[ht]
\centering
\caption{Detection results on MLT19 and language-wise performance. CRAFTS (paper) means that the result comes from the paper~\cite{baek2020character}. The result of CRAFTS$^{\ast}$ comes from the official ICDAR19-MLT website.}
\resizebox{0.8\linewidth}{!}{%
\begin{tabular}{@{}l cccc cccccccc@{}}
\toprule
Method & R & P & H & AP & Arabic & Latin & Chinese & Japanese & Korean & Bangla & Hindi \\ 
\midrule
PSENet~\cite{wang2019shape} & 59.59 & 73.52 & 65.83 & 52.73 & 43.96 & 65.77 & 38.47 & 34.47 & 51.73 & 34.04 & 47.19\\
RRPN~\cite{ma2018arbitrary} & 62.95 & 77.71 & 69.56 & 58.07 & 35.88 & 68.01 & 33.31 & 36.11 & 45.06 & 28.78 & 40.00\\
CRAFTS$^{\ast}$~\cite{baek2020character} & 62.73 & 81.42 & 70.86 & 56.63 & 43.97 & 72.49 & 37.20 & 42.10 & 54.05 & 38.50 & \textbf{53.50}\\ 
CRAFTS (paper)~\cite{baek2020character} & \underline{70.1}  & 81.7  & \underline{75.5} & -- & -- & -- & -- & -- & -- & -- & --\\
Single-head TextSpotter~\cite{liao2020mask}  & 61.76 & \underline{83.75} & 71.10 & 58.76 & 51.12 & \underline{73.56} & 40.41 & 41.22 & 56.54 & \underline{39.68} & 49.00 \\ 
Multiplexed TextSpotter~\cite{huang2021multiplexed} &63.16  & \textbf{85.53} & 72.66 & \underline{60.46} & \underline{51.75} & 73.55 & \underline{43.86} & \underline{42.43} & \underline{57.15} & \textbf{40.27} & \underline{51.95}\\ 
DBNet~\cite{liao2020real} & 64.0 & 78.3 & 70.4 & -- & -- & -- & -- & -- & -- & -- & --\\
DBNet++~\cite{liao2022real} & 65.4 & 78.6 & 71.4  & -- & -- & -- & -- & -- & -- & -- & --\\
\hline
ESTextSpotter-Polygon (Ours) & \textbf{75.5} & 83.37 & \textbf{79.24} & \textbf{72.52} & \textbf{52.00} & \textbf{77.34} & \textbf{48.20} & \textbf{48.42} & \textbf{63.56} & 38.26 & 50.83 \\ 
\bottomrule
\end{tabular}}
\label{tab:mlt19_task1}
\end{table*}

\subsection{Implementation Details}

We pre-train the model on a combination of Curved SynthText~\cite{liu2020abcnet}, ICDAR-MLT~\cite{nayef2017icdar2017}, and the corresponding datasets with $240$K iterations. The base learning rate is $1 \times 10^{-4}$ and reduced to $1 \times 10^{-5}$ at the $180$K-th iteration and $1 \times 10^{-6}$ at $210$K-th iteration. Then, the model is pre-trained on the Total-Text~\cite{ch2020total}, ICDAR 2013~\cite{karatzas2013icdar}, and ICDAR-MLT and fine-tuned on the corresponding real datasets. For Chinese and Vietnamese datasets, we follow the training strategies in previous works~\cite{m_Nguyen-etal-CVPR21,liu2021abcnetv2} to train the model. We use $N=100$ as the maximum number of predictions. The max length of recognition queries $T$ is $25$. The weight for the classification loss $\alpha_{c}$ is $2.0$. The weight of the $\ell_1$ loss is $5.0$ and of the GIoU loss is $2.0$.
The polygon loss weight $\alpha_{p}$ and the recognition loss weight $\alpha_{r}$ are both set to $1.0$. The focal loss parameters $\alpha$ and $\gamma$ are $0.25$ and $2.0$, respectively. The number of both encoder and decoder layers is 6. The inference speed is tested on a single NVIDIA GeForce RTX 3090.

The data augmentation strategies used 
are also kept the same as previous works~\cite{zhang2022text,liu2020abcnet,liu2021abcnetv2} as follows: 1) random resizing with the shorter size chosen from $640$ to $896$ pixels (with an interval of $32$), and the longest size is constrained within $1600$ pixels; 2) random cropping, which ensures that text is not being cut; 3) random rotation, which rotates the images with an angle in range of $[-45 ^ {\circ}$, $45 ^ {\circ}]$. For testing, we resize the shorter size of the image to $1000$ pixels while keeping the longest size of the image within $1824$ pixels.

\subsection{Comparison with State-of-the-Arts}

\paragraph{Multi-oriented Text.} We conduct experiments on ICDAR2015 and MSRD-TD500~\cite{yao2012detecting} to evaluate the robustness of our method for multi-oriented text. The detection results are presented in Table~\ref{tab:cp_sota_det}. Our method achieves the highest H-mean score of $91.0\%$ on the ICDAR2015 dataset, outperforming DeepSolo by $1.0\%$. On the MSRA-TD500 dataset, ESTextSpotter-Polygon achieves an accuracy of $89.5\%$. These results demonstrate the robustness of our method for detecting long, straight text. The end-to-end recognition results on the ICDAR2015 are shown in Table~\ref{tab:cp_sota_e2e}. Our method outperforms previous methods on all lexicon settings. Notably, in the strong lexicon setting, ESTextSpotter-Polygon achieves $87.5\%$ in terms of the Hmean, $2.3\%$ higher than the TESTR and TTS. In weak and generic lexicon, ESTextSpotter-Polygon outperforms the state-of-the-art implicit synergy method DeepSolo by $1.1\%$ and $1.2\%$, respectively. It demonstrates the effectiveness of the proposed explicit synergy.

\begin{table*}[!t]\small
  \centering
  \newcommand{\tabincell}[2]{\begin{tabular}{@{}#1@{}}#2\end{tabular}}
  \caption{End-to-end text spotting results on Total-Text, SCUT-CTW1500, ICDAR2015 and ReCTS. `None' means lexicon-free. `Full' indicates that we use all the words that appeared in the test set. `S', `W', and `G' represent recognition with `Strong', `Weak', and `Generic' lexicons, respectively.}
\resizebox{0.9\linewidth}{!}{%
\addtolength{\tabcolsep}{9pt}
  \begin{tabular}{@{}l cc cc ccc c c@{}}
    \toprule
    \multirow{2}*{Methods}  & \multicolumn{2}{c}{Total-Text} & \multicolumn{2}{c}{SCUT-CTW1500} &
    \multicolumn{3}{c}{ICDAR 2015 End-to-End}  &
    ReCTS & FPS \\
    \cmidrule(r){2-3} \cmidrule(r){4-5} \cmidrule(r){6-8} 
        & None & Full & None & Full  & S & W & G &1-NED & \\
    \midrule
    Mask TextSpotter \cite{liao2019mask}  &65.3&77.4  &--&-- &83.0&77.7&73.5 & 67.8  &--\\
    FOTS \cite{liu2018fots} &--&-- &21.1&39.7  &83.6&79.1&65.3 &50.8 &--\\
    TextDragon \cite{feng2019textdragon} &48.8&74.8 &39.7&72.4  &82.5&78.3&65.2 &-- &--\\
    Text Perceptron \cite{feng2019textdragon} &69.7 & 78.3 &57.0 & -- &80.5 &76.6 &65.1 & -- &--\\
    ABCNet \cite{liu2020abcnet} &64.2&75.7 &45.2&74.1  &--&--&-- &-- & 17.9 \\
    Mask TextSpotter v3 \cite{liao2020mask} &71.2&78.4 &--&--  &83.3&78.1&74.2 &-- &-- \\
    PGNet \cite{wang2021pgnet}  & 63.1 & --  & -- & --  & 83.3 & 78.3 & 63.5  & -- & 35.5 \\
    MANGO \cite{qiao2021mango}  & 72.9 & 83.6  & 58.9 & 78.7  & 81.8 & 78.9 & 67.3  & -- & 4.3 \\
    ABCNet v2 \cite{liu2021abcnetv2}  & 70.4 & 78.1  & 57.5 & 77.2  & 82.7 & 78.5 & 73.0  & 62.7 & 10.0 \\
    PAN++ \cite{liu2021abcnetv2}  & 68.6 & 78.6  & -- & --  & 82.7 & 78.2 & 69.2  & -- & 21.1 \\
    Boundary TextSpotter'22 \cite{luboundary}  & 66.2 & 78.4  & 46.1 & 73.0  & 82.5 & 77.4 & 71.7  & -- & 13.4 \\
    SwinTextSpotter \cite{huang2022swintextspotter}  & 74.3 & 84.1  & 51.8 & 77.0  & 83.9 & 77.3 & 70.5  & 72.5 & 1.0 \\
    SRSTS \cite{wu2022decoupling}  & 78.8 & 86.3  & -- & --  & 85.6 & 81.7 & 74.5  & -- & 18.7 \\
    TPSNet \cite{wang2022tpsnet} & 78.5 & 84.1 & \underline{60.5} & 80.1  & -- & -- & --  & -- &-- \\
    GLASS \cite{ronen2022glass}  & \underline{79.9} & 86.2  & -- & --  & 84.7 & 80.1 & 76.3  & -- & 3.0 \\
    TESTR \cite{zhang2022text}  & 73.3 & 83.9  & 56.0 & \underline{81.5}  & 85.2 & 79.4 & 73.6  & -- & 5.3 \\
    TTS \cite{kittenplon2022towards}  & 78.2 & 86.3  & -- & --  & 85.2 & 81.7 & \underline{77.4}  & -- &-- \\
    ABINet++ \cite{fang2022abinet++} & 77.6 & 84.5 & 60.2 & 80.3 & 84.1 & 80.4 & 75.4 & \underline{76.5} & 10.6 \\
    DeepSolo \cite{ye2022deepsolo}  & 79.7 & \underline{87.0}  & 64.2 & 81.4  & \underline{86.8} & \underline{81.9} & 76.9  & -- & 17.0 \\
    \hline
    ESTextSpotter-Bezier (Ours)  & 79.7 & 86.0 & \textbf{66.0}  & 83.6  & \textbf{88.1} & 82.9 & 77.9  & - & 4.3 \\
    ESTextSpotter-Polygon (Ours)  & \textbf{80.8} & \textbf{87.1} & 64.9  & \textbf{83.9}  & 87.5 & \textbf{83.0} & \textbf{78.1}  & \textbf{78.1} & 4.3 \\
    \bottomrule
  \end{tabular}
}
\label{tab:cp_sota_e2e}
\end{table*}

\paragraph{Arbitrarily-Shaped Text.} We test our method on two arbitrarily-shaped text benchmarks (Total-Text and CTW1500) to verify the generalization ability of our approach for arbitrarily-shaped scene text spotting. For text detection task, as shown in Table~\ref{tab:cp_sota_det}, ESTextSpotter-Polygon outperforms the previous state-of-the-art model with $90.0\%$ in terms of the H-mean metric on Total-Text dataset, $2.6\%$ higher than the DeepSolo. On the CTW1500 dataset, our method also achieves $90.0\%$, which also significantly outperforms previous methods. The end-to-end scene text spotting results are shown in Table~\ref{tab:cp_sota_e2e}, ESTextSpotter-Polygon significantly surpasses the TTS by a large margin ($2.6\%$ without lexicon and $0.8\%$ on ``Full'' lexicon) on TotalText. On the CTW1500 dataset, ESTextSpotter-Polygon outperforms all previous best models by $0.7\%$ without lexicon and $2.4\%$ on `Full' lexicon. On CTW1500, ESTextSpotter-Bezier presents $66.0\%$ performance without lexicon.
From Tables~\ref{tab:cp_sota_det} to~\ref{tab:cp_sota_e2e}, it can be seen that our method consistently achieves the best results for text detection and text spotting.

\begin{table}[!t]
\centering
\caption{End-to-end text spotting results on VinText. ABCNet+D means adding the methods proposed in \cite{m_Nguyen-etal-CVPR21} to ABCNet. The same to Mask Textspotter v3+D.}
\resizebox{0.55\linewidth}{!}{
\begin{tabular}{@{}l c@{}}
\hline
Method & H-mean                       \\ \hline 
ABCNet\cite{liu2020abcnet}                  & 54.2                     \\ 
ABCNet+D\cite{m_Nguyen-etal-CVPR21}     & 57.4                     \\ 
Mask Textspotter v3\cite{m_Nguyen-etal-CVPR21}     & 53.4                     \\ 
Mask Textspotter v3+D\cite{m_Nguyen-etal-CVPR21}     & 68.5                     \\ 
SwinTextSpotter\cite{huang2022swintextspotter}                    & \underline{71.1}                        \\  \midrule
ESTextSpotter-Polygon (Ours)                    & \textbf{73.6}                        \\  \hline
\end{tabular}}
\label{tab:VinText}
\end{table}

\begin{table}[!t]\small
\centering

\addtolength{\tabcolsep}{2pt}
\caption{End-to-end text spotting results and detection results on HUST-ART.}
\resizebox{0.8\linewidth}{!}{%
    \begin{tabular}{@{}l ccc c@{}}
    \toprule
    \multirow{2}{*}{Method} & \multicolumn{3}{c}{Detection} & \multirow{2}{*}{E2E} \\ \cline{2-4}
                            & P        & R        & H        &                        \\ 
    \midrule
    DB\cite{liao2020real}                  & 95.31   & 74.62  & 83.71     & --                    \\ 
    DCLNet\cite{bi2021disentangled}                 &  93.82  & 77.47 &  84.86   & --    \\
    PAN++\cite{wang2021pan++}                  &  93.38 &  30.06 & 45.48     & 30.31                    \\ 
    MaskTextSpotter v3\cite{wang2021pan++}               & 88.31  & 80.82 & \underline{84.40}     & \underline{71.23}                    \\ \hline
    ESTextSpotter-Polygon (Ours)                   & \textbf{96.05}     & \textbf{82.79}     & \textbf{88.93}     & \textbf{77.55}                   \\
    \bottomrule
    \end{tabular}}%
    \vspace{-5pt}
\label{tab:hust_art}
\end{table}

\begin{figure*}[!t]
    \centering
    \subcaptionbox{\small Total-Text}
    {\includegraphics[width=5.6cm,height=3.0cm]{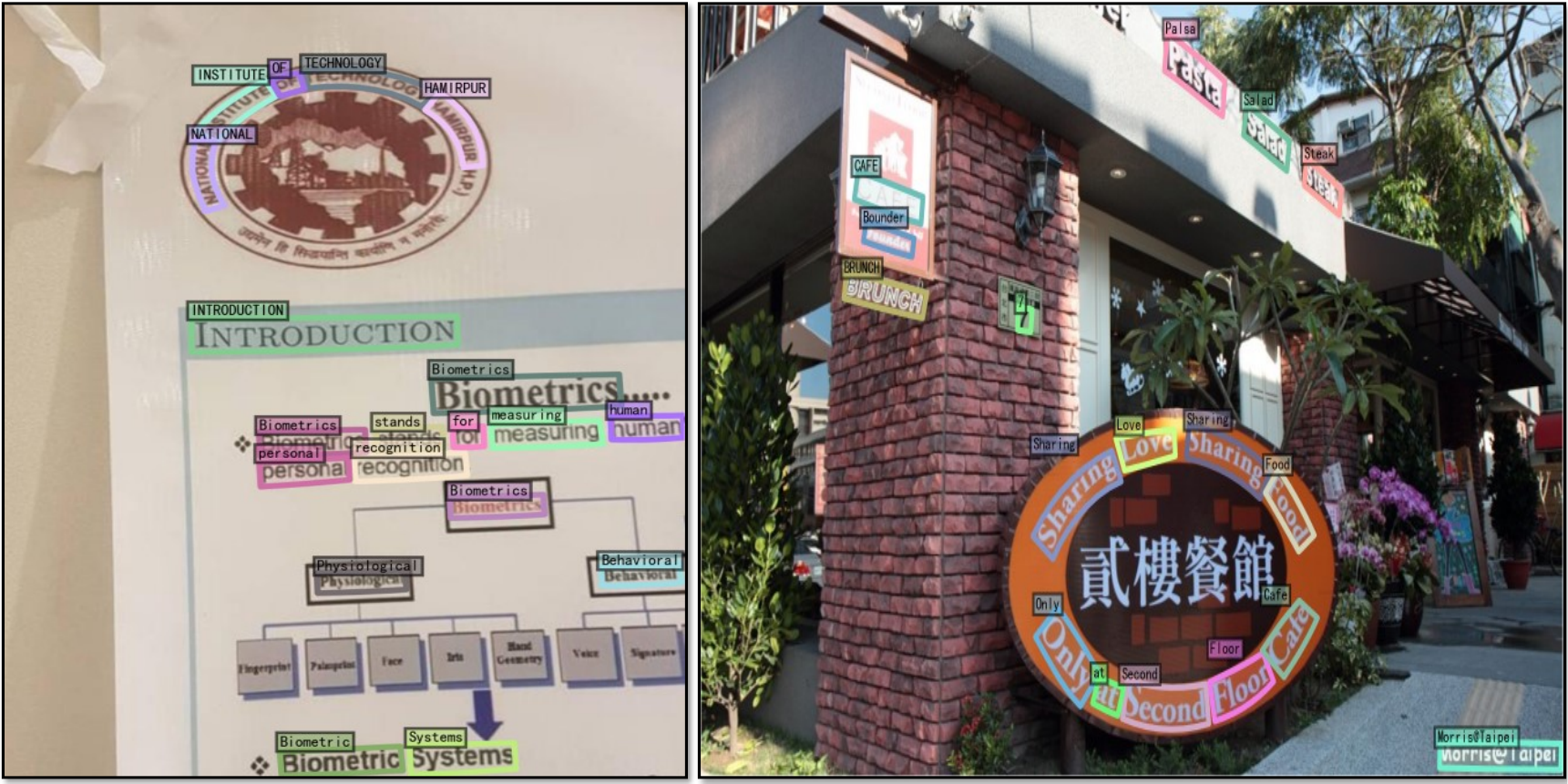}\label{fig:vis_tt}}
    \subcaptionbox{\small SCUT-CTW1500}
    {\includegraphics[width=5.6cm,height=3.0cm]{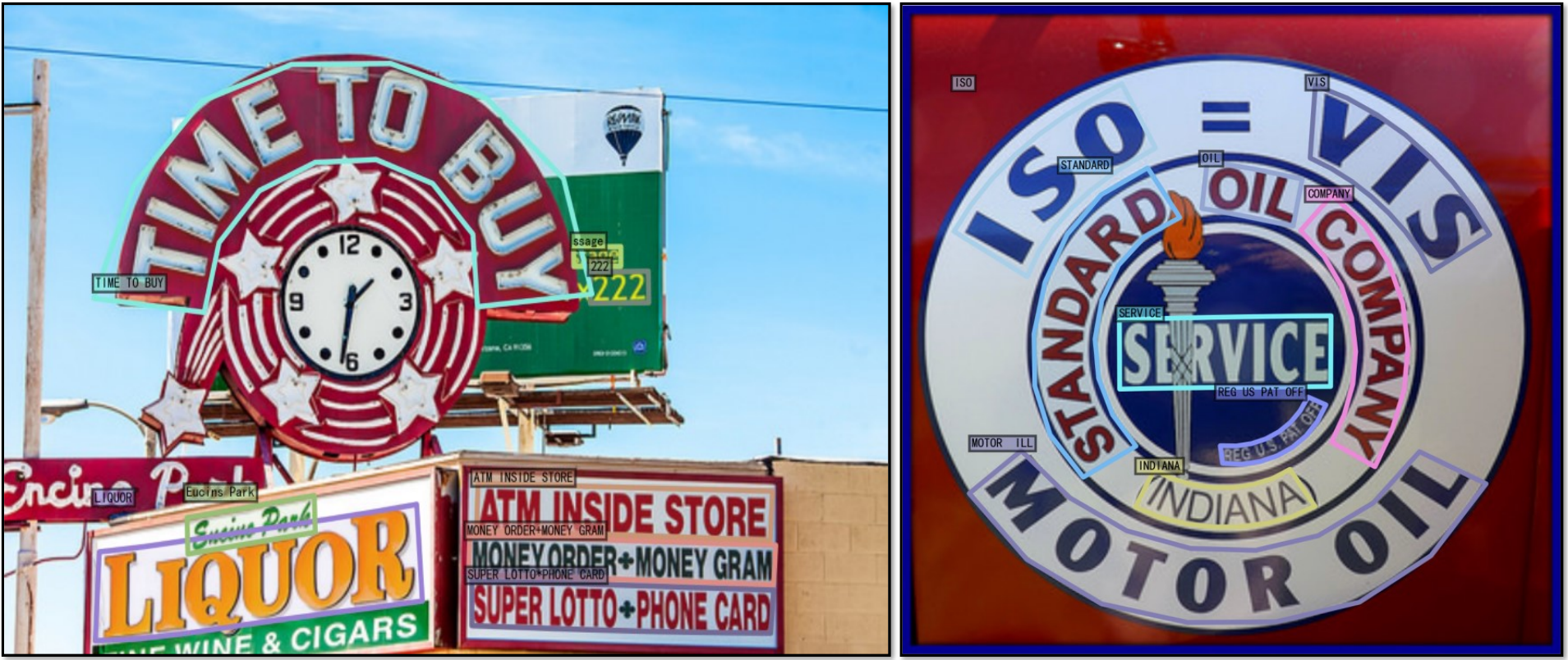}\label{fig:vis_ctw1500}}
    \subcaptionbox{\small VinText}
    {\includegraphics[width=5.6cm,height=3.0cm]{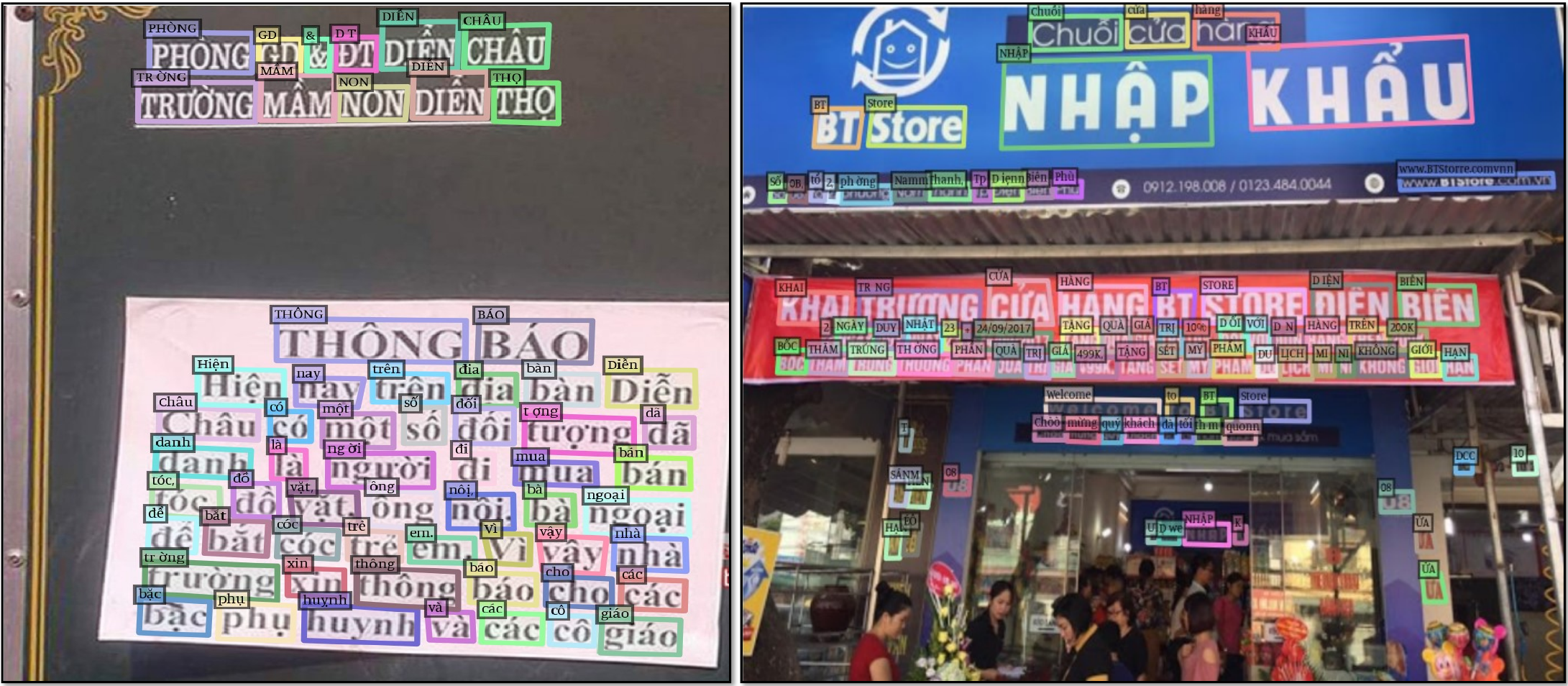}\label{fig:vis_vintext}}
    \subcaptionbox{\small ReCTS}
    {\includegraphics[width=5.6cm,height=3.0cm]{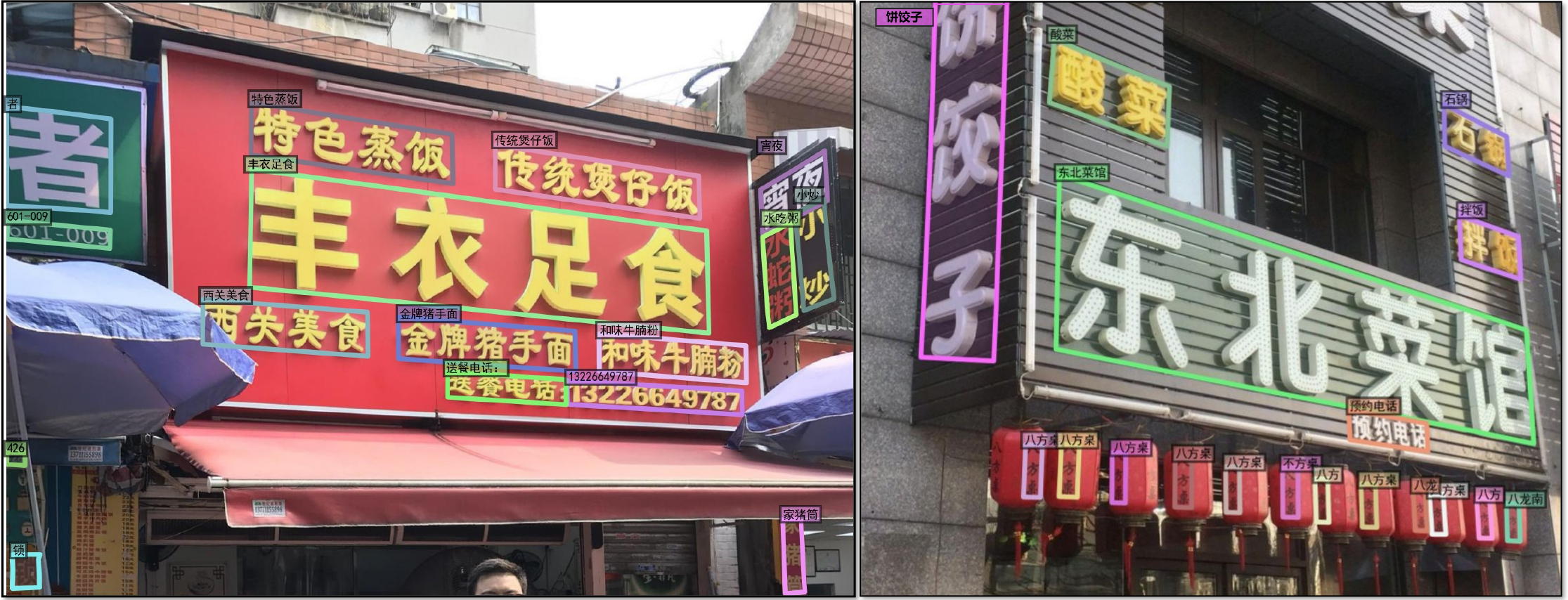}\label{fig:vis_rects}}
    \subcaptionbox{\small ICDAR2015}
    {\includegraphics[width=5.6cm,height=3.0cm]{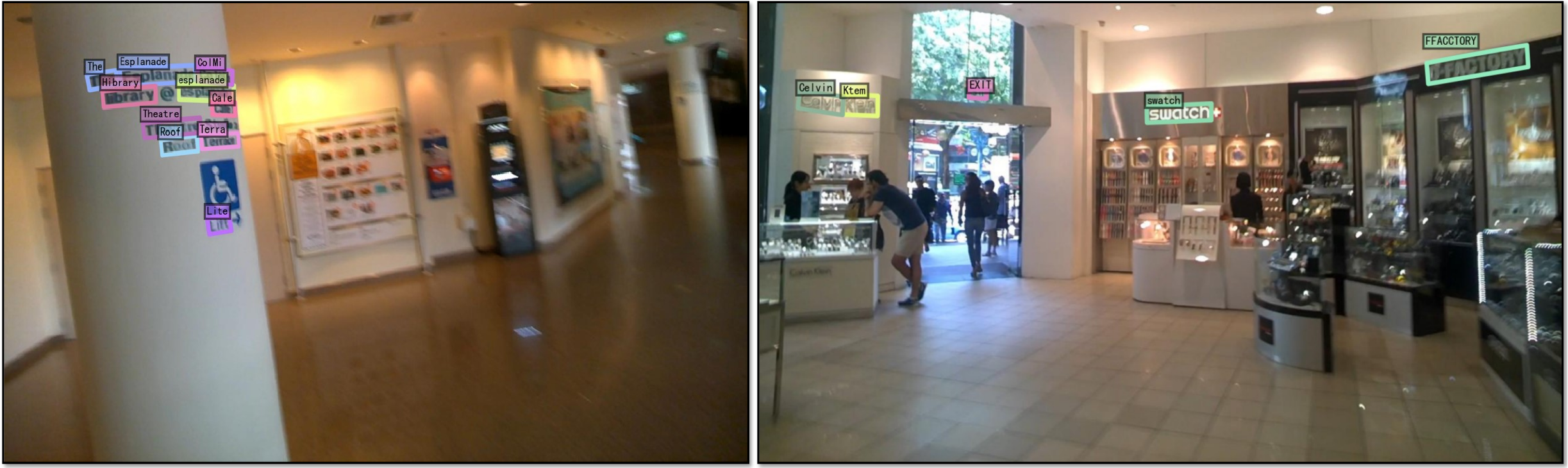}\label{fig:vis_ic15}}
    \subcaptionbox{\small HUST-ART}
    {\includegraphics[width=5.6cm,height=3.0cm]{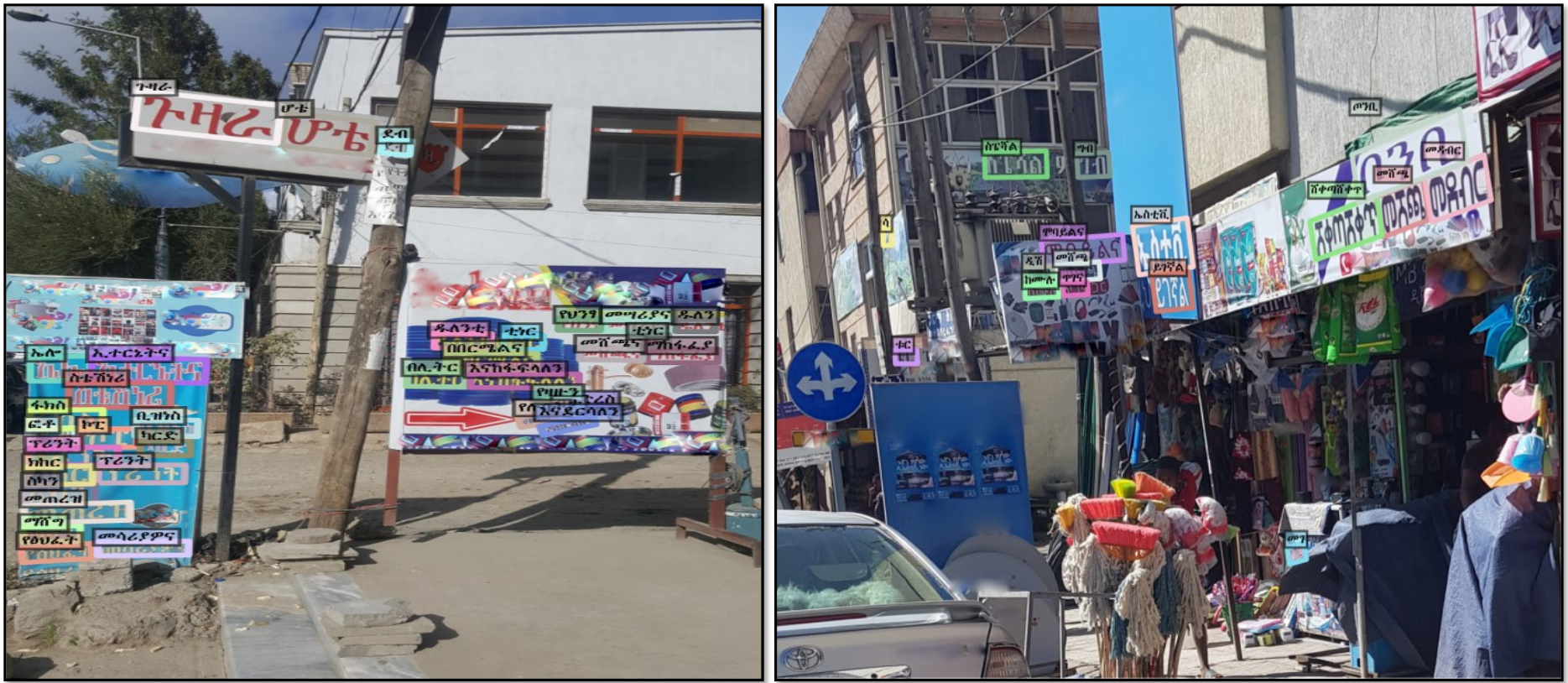}\label{fig:vis_hust}}
    \caption{Visualization results of our ESTextSpotter-Polygon on different datasets. Best viewed in screen.}
    \label{fig:vis_results}
\end{figure*}

\paragraph{Multilingual Text.} We further evaluate ESTextSpotter-Polygon using multilingual datasets. The results for the ReCTS can be found in Tables~\ref{tab:cp_sota_det} and~\ref{tab:cp_sota_e2e}, which showcase ESTextSpotter's superior performance over the state-of-the-art method in both detection and text spotting. Notably, our method outperforms ABINet++, a method leveraging iterative language modeling for text spotting, by $1.6\%$ in terms of text spotting performance. For VinText, the result is shown in Table~\ref{tab:VinText}, from which we can see ESTextSpotter-Polygon outperforms the SwinTextSpotter by $2.5\%$. Note that ABCNet+D and Mask TextSpotter v3+D mean using the dictionary to train the model, which is not used by our method. For the HUST-ART~\cite{dikubab2022comprehensive}, shown in Table~\ref{tab:hust_art}, our method achieves the best performance on both detection and end-to-end recognition, significantly outperforming MaskTextSpotter v3. For the well-known multilingual benchmark ICDAR2019-MLT, the detection results and language-wise H-mean are shown in Table~\ref{tab:mlt19_task1}. Our method achieves the best performance in all languages except Bangla and Hindi. We provide some qualitative results in Figure~\ref{fig:vis_results}.

\begin{table*}[!t]\small
    \centering
    \caption{\small Ablation studies on Total-Text. ``None'' represents lexicon-free. IS means implicit interaction within the decoder. ES means explicit interaction within the decoder. VLC means the vision-language communication module in task-aware decoder. TAQI is the task-aware query initialization. TADN means the task-aware denoising training. REM means receptive enhancement module. }
\resizebox{.95\linewidth}{!}{
\setlength{\tabcolsep}{14pt}
    \begin{tabular}{@{}l c c c c c c ccc c@{}}
    \toprule
    \multirow{2}{*}{Method} & \multirow{2}{*}{IS} & \multirow{2}{*}{ES} & \multirow{2}{*}{TAQI} & \multirow{2}{*}{VLC}  & \multirow{2}{*}{REM} & \multirow{2}{*}{TADN} & \multicolumn{3}{c}{Detection} & \multicolumn{1}{c}{End-to-End} \\ \cmidrule(lr){8-10}
     &  &  &   & & & & P & R & F & None  \\ 
     \midrule
    Baseline &  &  &  &  &  &  & 89.5 & 84.8 & 87.1 & 68.2 \\ 
    Baseline & \checkmark &  &  &  &  &  & 87.0 & 81.4 & 84.1 & 70.2 \\ 
    Baseline &  & \checkmark &  &  &  &  & 90.6 & 85.0 & 87.7 & 70.4 \\ 
    Baseline &  & \checkmark & \checkmark &  &  &  & 90.3 & \textbf{86.5} & 88.1& 70.7 \\ 
    Baseline &  & \checkmark & \checkmark & \checkmark &  &  & 90.4 & 86.0 & \textbf{88.3} & 72.0 \\
    Baseline &  & \checkmark & \checkmark & \checkmark & \checkmark &  & 90.3 & 86.2 & 88.2 & 72.7 \\
    ESTextSpotter-Polygon &  & \checkmark & \checkmark & \checkmark & \checkmark & \checkmark & \textbf{90.7} & 85.3 & 87.9 & \textbf{73.8} \\
    \bottomrule
    \end{tabular}
}
\label{tab:ablation}
\vspace{-0.5em}
\end{table*}

\subsection{Ablation Studies}
\label{sec:ablation}
We conduct ablation studies on the Total-Text to investigate the impact of different components in ESTextSpotter-Polygon. For text spotting, the results contain a bias caused by randomness. Our experiments show that the pre-trained model had a bias of $0.2\%$, while the finetuned model had a bias of $0.5\%$ in the end-to-end text spotting results. In the ablation studies, we use the pre-trained model to evaluate the results.

\paragraph{Comparison between implicit synergy and explicit synergy.} To verify that our proposed explicit synergy achieves better synergy than the previous implicit synergy, we conduct experiments to validate that explicit synergy achieves better synergy compared to implicit synergy. The explicit synergy develops task-aware queries to conduct explicit interaction between detection and recognition within the decoder by modeling two distinct feature patterns for each task and interacting with each other. In contrast, the previous implicit synergy solely relies on shared features that overlook the divergent feature requirements of two tasks and lack explicit modules to ensure the interaction, resulting in limited synergy.
The results are shown in Table~\ref{tab:ablation}. Although implicit synergy can improve text spotting performance, they often lead to a degradation in detection. In contrast, our proposed explicit synergy improves both detection and spotting performance, demonstrating its superior synergy.

\paragraph{Task-aware query initialization (TAQI) and vision-language communication module (VLC).} The results shown in Table~\ref{tab:ablation} demonstrate that the TAQI could lead to improvements of $0.4\%$ and $0.3\%$ for detection and end-to-end scene text spotting, respectively. Moreover, the VLC could further enhance the performance by $0.2\%$ and $1.3\%$ for detection and end-to-end scene text spotting, respectively. 
It demonstrates that conducting TAQI and VLC can promote stable explicit synergy and greatly enhance performance for text detection and spotting.

\paragraph{Receptive Enhancement Module (REM) and Task-aware denoising training (TADN).} The results presented in Table~\ref{tab:ablation} demonstrate that the REM can result in a $0.7\%$ improvement in text spotting performance. Furthermore, TADN was employed, leading to an additional improvement of $1.1\%$ in text spotting performance. Notably, TADN is more focused on text spotting rather than detection, as opposed to previous denoising training methods~\cite{li2022dn,zhang2022dino}.

\subsection{Visualization and Analysis}

The visualization results are shown in Figure~\ref{fig:vis_results}. It can be observed that our method can accurately detect and recognize the texts. Notably, ESTextSpotter-Polygon is effective in recognizing horizontally arranged vertical text (Figure~\ref{fig:vis_results}(d)). This is due to the sufficient interaction between text detection and recognition using our explicit synergy. It provides the recognition queries with the text orientation, which helps determine the reading order of characters.  

\section{Conclusion}

In this paper, we present a simple yet effective Transformer with explicit synergy for text spotting. Previous implicit synergy can not fully realize the potential of two tasks. To address this issue, our approach explores explicit synergy to allow task-aware queries to explicitly model the discriminative and interactive features between text detection and recognition within a single decoder. Additionally, our proposed vision-language communication module enables task-aware queries to conduct interactions from a cross-modal perspective, thereby unleashing the potential of both text detection and recognition. Extensive experiments on a wide range of various benchmarks, including multi-oriented, arbitrarily-shaped, and multilingual datasets, consistently demonstrate that our method outperforms previous state-of-the-art approaches by significant margins. We hope our method can inspire further investigation on the explicit synergy in text spotting area.

\noindent \paragraph{Acknowledgement} This research is supported in part by NSFC (Grant No.: 61936003) and Zhuhai Industry Core and Key Technology Research Project (no. 2220004002350).

{\small
\bibliographystyle{ieee_fullname}
\bibliography{ests}
}

\end{document}